\documentclass[11pt]{article}
\usepackage[margin=1in]{geometry}
\usepackage{latexsym}
\usepackage{graphicx}
\usepackage{mathptmx}
\usepackage[T1]{fontenc}

\usepackage{amsmath}
\usepackage{amsfonts}
\usepackage{amssymb}
\usepackage{amsbsy}
\usepackage{comment}
\usepackage{multirow}
\usepackage{subcaption}

\usepackage[numbers]{natbib}

    \newcommand{\shortcite}[1]{\cite{#1}}
    \newcommand{\citeN}[1]{\citet{#1}}
    \newcommand{\shortciteN}[1]{\citet{#1}}

\usepackage[colorlinks=true,urlcolor=blue,citecolor=black,anchorcolor=black,linkcolor=black]{hyperref}

\begin{document}

\title{On the Potential of Graph Neural Networks as Metamodels for Supply Chain Optimization: Dataset, Architectures, and Directions}

\author{Tushar Lone and Neha Karanjkar\\
School of Mathematics and Computer Science,\\
Indian Institute of Technology Goa, Ponda, Goa, INDIA}

\date{}

\maketitle

\begin{abstract}
Graph Neural Networks (GNNs) have emerged as a powerful, differentiable class of learning models for graph-structured systems. Their ability to generalize across topologies opens the prospect of a surrogate for combined structural and parametric optimization, which classical metamodels cannot offer. Supply chains are a natural target, yet the use of GNN surrogates for supply chain problems is largely unexplored. This paper lays the foundation, presents initial steps, and discusses key research directions. As a foundation, we formulate the problem and create a large public training dataset of programmatically generated supply chain graphs with input parameters and steady-state performance metrics obtained using our SupplyNetPy simulation library. As initial steps, we explore GNN architectures that work well as surrogates for node- and network-level predictions, and analyze their accuracy-compute trade-off against simulation. Most importantly, we outline the exciting directions this opens, namely gradient-based optimization over topology, fast design-space exploration, and sensitivity analysis.
\end{abstract}

\section{Introduction}
\label{sec:intro}

Modern supply chains (SCs) are large, complex networks of suppliers, factories, warehouses, and retailers that require analysis and optimization to meet demand, minimize costs, and increase profit. Analytical optimization methods are restrictive and do not capture SCs' intricate, stochastic behavior. Simulation methods model this complexity but make simulation-based optimization (SBO) computationally expensive. SBO is a black-box problem where gradients are unavailable, local optimizers cannot be applied directly, and global optimizers require many costly evaluations~\shortcite{pedrielli2019metamodel,xie2019metamodel}.

A metamodel approximates the simulation's input-output behavior, trading some accuracy for speed~\shortcite{cen2022enhanced,amaranath2023causal}. Trained on simulation data, it provides a smooth surrogate that yields gradient information and supports local optimization, letting the optimizer search more efficiently than the full simulation.
Selecting a metamodel is therefore important. Common choices include Gaussian Process Regression (GPR), Polynomial Response Surfaces (PRS), and Multi-Layer Perceptrons (MLPs). In prior work~\shortcite{lone2023open} we found that GPR (Kriging) and standard neural networks (NN) fail to capture topological information and thus cannot predict performance across arbitrary supply chain networks (SCNs).
We therefore explore Graph Neural Networks (GNNs) as metamodels. GNNs learn from graph structure by aggregating node and edge information and are differentiable end-to-end, making them well suited as differentiable metamodels for SC optimization~\shortcite{schuetz2022combinatorial,lee2025diffim}. SCs are naturally graph-structured, with facilities as nodes and transport or information links as edges~\shortcite{wasi2024supplygraph}. We pose the following questions:

\textit{(Q1)} Can GNNs serve as metamodels in SBO of SCs, and what architecture suits this task?

\textit{(Q2)} How can we generate training data for the GNN metamodel?

\textit{(Q3)} Do GNN metamodels generalize to SCNs not seen during training?

\textit{(Q4)} Are GNN metamodels computationally efficient relative to running the simulation directly?

\textit{(Q5)} Can we leverage GNNs' differentiability for optimization and design-space exploration?

To address these questions, we design a custom GNN architecture that incorporates node and edge features to predict node-level performance metrics, and we use the SupplyNetPy library~\cite{supplynetpy_github} to generate a large, diverse synthetic dataset of arbitrary two-echelon SCNs without relying on proprietary data. We evaluate generalization on held-out larger networks and assess trade-offs between model complexity, prediction accuracy, and computational cost. Q1, Q3, and Q4 are addressed empirically, Q2 by the dataset contribution, while Q5 is a forward-looking direction we motivate rather than demonstrate, so this paper lays the foundation and takes initial steps toward GNN-based SC optimization.

\textit{Contributions.}
The central demonstrated contribution of this paper is to establish that a single GNN metamodel predicts node- and network-level performance across SCNs of differing shape and size, and generalizes to networks far larger than those seen in training. The contributions are:

\textit{(C1) A custom, structure-aware GNN metamodel.} An edge-conditioned, node-type-aware message-passing layer that predicts node- and network-level performance and demonstrates generalization from a trained model to networks more than twenty times larger than those seen during training (Q1 and Q3).

\textit{(C2) A synthetic dataset that enables this study.} A large-scale dataset of 100,000 diverse, arbitrarily configured two-echelon SCNs generated with SupplyNetPy, made public to support future research (Q2).

\textit{(C3) Design decisions and accuracy-compute trade-off.} A systematic study of how dataset size and hyperparameters affect prediction accuracy and training time, with a preliminary comparison against full simulation that points to the GNN surrogate's computational advantage inside optimization loops (Q4).

\textit{(C4) Research agenda for differentiable, structure-aware SC optimization.} An explicit case for why GNN metamodels are uniquely positioned to enable joint parameter and topology optimization via gradient-based methods, design-space exploration, and sensitivity analysis through the trained metamodel (Q5).

\textit{Paper organization.}
Section~\ref{sec:lit_review} reviews related work. Section~\ref{sec:data} introduces SupplyNetPy, the experimental design, modeling assumptions, and datasets. Section~\ref{sec:gnn_arch} describes the custom message-passing GNN and evaluates accuracy and generalization. Section~\ref{sec:tradeoffs} examines computational trade-offs. Section~\ref{sec:conclusion} concludes and discusses future directions.
\section{Related Work}
\label{sec:lit_review}

To understand the current state of research on the application of GNN-based metamodels for SBO in SCs, we pose the following questions.

\textit{(L1)} Have GNNs been applied as metamodels in simulation-based studies?

\textit{(L2)} What learning-based metamodels have been applied to SC analysis and optimization?

\textit{(L3)} Have GNNs been applied to SC problems, and what prediction tasks and architectures are used?

\textit{(L4)} Have GNN architectures been systematically explored to assess accuracy and computational cost trade-offs?

\textit{(L5)} Has sensitivity analysis been performed using GNNs for SC optimization?

The review is conducted on Google Scholar with keyword combinations drawn from GNN terminology and SC problem domains, including terms: \textit{graph neural network}, \textit{supply chain}, \textit{inventory}, \textit{simulation}, \textit{optimization}, and \textit{metamodel}.
This is a focused narrative review organized around the questions rather than a formal systematic review, and within each theme we cite a representative subset of the works identified.

\subsection{GNNs as Metamodels in Simulation-Based Studies}
\label{subsec:gnn_as_metamodel}
Many studies have applied GNNs to simulation-based research across different domains, including solid and fluid mechanics, materials design, molecular and atomistic simulations, social networks, financial systems, traffic and transportation, SCs, and logistics and routing. To address Question \textit{L1}, we review relevant studies that apply GNN metamodels in the SC-relevant domain.

\citeN{cen2022enhanced} propose a message-passing neural network (MPNN) metamodel for SBO of stochastic activity networks, which model manufacturing order-assembly tasks as directed activity graphs.
Trained on simulation data, it predicts the expected task completion time from both the graph structure and continuous service-rate parameters, making the graph structure itself a variable input and overcoming the fixed-length vector limitation of traditional metamodels.
\citeN{cen2023efficient} extend this metamodel to optimization with a hybrid algorithm that combines Monte Carlo tree search with automatic differentiation through the network, jointly optimizing discrete design variables such as edge removal and continuous service rates and computing gradients by backpropagation to guide the search by sensitivity.
The key takeaway from these studies is that GNNs can be effective metamodels for graph-structured input data.

\subsection{Learning-Based Metamodels for Supply Chain Analysis and Optimization}
\label{subsec:metamodels_in_sc}
Addressing Question \textit{L2}, metamodels are widely used in SBO to approximate the simulation's input-output behavior and increase the efficiency of SBO methods.
For SC problems (such as inventory management and manufacturing systems) and queueing systems, learning-based metamodels such as NNs, GPR, and PRS have been applied. Below we review relevant studies that address these problems using metamodels.

In our previous work~\shortcite{lone2023open}, we evaluate NN and Kriging metamodels on an 8-parameter, 5-node inventory optimization problem.
These are effective for a fixed SCN but fail to capture its network structure and therefore cannot be applied to arbitrary SCs.
\citeN{pedrielli2019metamodel} apply polynomial and quantile regression metamodels, trained on simulation data from a three-stage manufacturing system, to support real-time job-release control decisions.
\shortciteN{xie2019metamodel} combine a stochastic Kriging metamodel with a global sensitivity analysis framework based on functional analysis of variance on a biopharma inventory simulation, identifying which inputs contribute most to output uncertainty as a form of metamodel-assisted sensitivity analysis.
\shortciteN{amaranath2023causal} introduce modular dynamic Bayesian networks as a metamodel for stochastic M/M/1 queue simulations, supporting probabilistic, interventional, and inverse queries, where inverse queries identify the parameters that maximize a performance objective.

To summarize, existing learning-based metamodels for SCs treat the simulation as a mapping from a fixed-length input vector to a scalar output.
They do not account for the SC graph topology as an input.

\subsection{Applications of GNNs in Supply Chains}
\label{subsec:gnn_in_sc}
Addressing Question \textit{L3}, a growing body of work applies GNNs to specific SC problems, primarily using real or domain-specific datasets.
Key architectures include GCNs~\cite{kipf2016semi}, GATs, GraphSAGE~\shortcite{hamilton2017inductive}, and MPNN~\shortcite{gilmer2017neural}, with GraphSAGE applied to SC visibility tasks~\cite{zheng2025machine,zheng2025analytics}. Furthermore, temporal GNNs (incorporating long short-term memory or Transformer modules) are used for time-series tasks (demand forecasting).
A recent benchmark by \shortciteN{wasi2024supplygraph}, covering GCN, GAT, and GIN across six SC analytics tasks using the real-world SupplyGraph dataset, shows that GNNs consistently outperform traditional machine learning and deep learning methods.

Several studies combine GNNs with reinforcement learning (RL) for multi-echelon inventory optimization. 
\shortciteN{ziegner2025iterative} encode SC state via a GNN to drive multi-agent RL policies across echelons, \citeN{feng2025joint} apply a GAT with multi-agent RL for joint scheduling and replenishment, and \citeN{guo2024research} use GNN-based state encoding for cost optimization across multiple echelons.
\shortciteN{ahn2024gnn} use a GAT for probabilistic node-level regression of supply and inventory levels from real historical data, capturing multi-tier network dependencies.
\shortciteN{weng2022semi} apply a GCN for semi-supervised classification of auto parts inventory.
Across these applications, GNNs serve predominantly as predictors for forecasting, classification, and reinforcement-learning control on real or domain-specific datasets, rather than as simulation metamodels that predict performance over arbitrary synthetic topologies.
Regarding Question \textit{L4}, to the best of our knowledge none of these studies systematically compares GNN architectures for the accuracy-compute trade-off in a metamodeling setting.

\subsection{GNN Differentiability for Optimization and Sensitivity Analysis}
\label{subsec:gnn_diff}
Addressing Question \textit{L5}, and looking beyond SCs, a growing line of work exploits the end-to-end differentiability of GNNs for optimization and sensitivity analysis on graphs.
GNNs have been trained as differentiable solvers and relaxations for combinatorial optimization on graphs, scaling to millions of variables~\shortcite{schuetz2022combinatorial,karalias2020erdos}, and for structure learning of directed acyclic graphs~\shortcite{yu2019dag}, though simple greedy heuristics can still outperform such solvers on some sparse-graph problems~\cite{angelini2023modern}.
Most relevant to our future direction, \shortciteN{lee2025diffim} train a GNN as a differentiable surrogate for influence propagation and interpret each edge gradient as its sensitivity, illustrating more broadly that GNN differentiability is a viable mechanism for gradient-based optimization and sensitivity analysis on graph-structured problems.

In summary, to the best of our knowledge no prior work applies GNNs as metamodels for SBO in SCs, the closest being studies on stochastic activity networks~\cite{cen2022enhanced,cen2023efficient}, and existing learning-based SC metamodels (GPR, NN, polynomial) assume a fixed topology that cannot generalize across network sizes or configurations. GNNs have been applied to SC tasks on real datasets with strong empirical gains, yet no study systematically evaluates their architecture trade-offs for metamodeling, and their differentiability has been exploited for sensitivity-guided optimization only in adjacent domains~\cite{cen2023efficient}.
\section{Training Dataset Generation}
\label{sec:data}

This section addresses Q2 from the introduction: how to generate training data for the GNN metamodel.
Real, large-scale SC data is scarce, since it is difficult to gather across many nodes, is often commercially sensitive, and typically covers a single aspect at one node, such as demand forecasting or logistics, rather than inventory levels, transport costs, and demand patterns across nodes.
The SupplyGraph benchmark~\shortcite{wasi2024supplygraph} is a notable exception, but its single fixed topology and temporal-forecasting focus do not provide the diverse graph topologies of varying size and configuration that training a GNN metamodel for SBO requires.
To address this gap, we use SupplyNetPy, a simulation library we developed, to generate synthetic SC data that closely represents real SC behavior.
The following subsections introduce the library and describe the workflow for data generation.

\subsection{Overview of SupplyNetPy}
\label{subsec:lib_intro}

SupplyNetPy~\cite{supplynetpy_github} is an open-source, component-based SC simulation library built on Python's SimPy.
It supports arbitrary multi-echelon network topologies, multiple replenishment policies, perishable inventory, node and link disruptions, and stochastic demand and lead times, and all of its components are extensible through inheritance.
Users describe an SC as a graph with node and link attributes, and the library performs the discrete-event simulation (DES) and produces node- and network-level performance reports.
It plays a critical role in this work because it lets us programmatically generate arbitrary SCN topologies with randomized parameters and simulate them to obtain ground-truth performance metrics, eliminating the dependence on proprietary software or scarce real-world datasets for training data generation.
For more details, visit the library's \href{https://supplychainsimulation.github.io/SupplyNetPy}{documentation} and \href{https://github.com/SupplyChainSimulation/SupplyNetPy}{GitHub}~\cite{supplynetpy_github} for usage examples, and the publicly available training \href{https://github.com/SupplyChainSimulation/Supply_chain_training_datasets}{datasets}~\cite{supplynetpy_data}, including those generated with different parameter configurations.

\subsection{Workflow for Data Generation}
\label{subsec:experiment_design}

We model inventory flow in a two-echelon SCN comprising distributors, retailers, and a single external infinite supply.
In particular, we model inventory capacity, reorder level, and holding costs at every distributor and retailer.
The dataset consists of such arbitrary two-echelon SCNs, each containing a single non-perishable product, a random number of nodes, and random connections between them.
Each distributor and retailer holds inventory and follows a min-max $(s,S)$ replenishment policy, reordering up to the order-up-to level $S$ whenever its inventory falls to the reorder level $s$.

Each data point in the dataset corresponds to an SCN with a specific configuration of nodes and edges. The simulation produces five per-node performance metrics, namely inventory carry cost (inventory holding cost over time), inventory spend cost (replenishment expenditure from upstream), transport cost, revenue, and profit (revenue minus the three costs). Network-level metrics are obtained by summing over all nodes. The node, edge, and network-level parameters and the performance metrics in the dataset are those we identified as shared across various SC applications, and SC problems in recent SC modeling and simulation studies, building on our earlier work~\shortcite{lone2023open,lone2024development}, and are tabulated in a \href{https://github.com/SupplyChainSimulation/InventOpt/tree/main/review}{public repository}~\cite{InventOpt}.
The following steps describe the process of generating a single data point, which is repeated to produce the full dataset.

First, a single fixed supplier with infinite inventory is created to supply all distributors. The supplier lies outside the two echelons and serves as an external source that ensures distributors can always replenish. Next, $K$ distributors and $N$ retailers are created, with $K$ sampled uniformly from 1 to 5 and $N$ from $K$ to 10, each following a min-max replenishment policy with inventory capacity and reorder level sampled uniformly at random, and the distributors' levels set sufficient to serve their connected retailers. Each retailer is then assigned a daily demand arrival rate sampled uniformly and modeled as a retailer attribute rather than a separate demand node, and the order quantity is kept constant (deterministic demand) so that a single simulation run suffices per data point. Finally, each distributor connects to $M$ retailers chosen at random, with $M$ sampled uniformly from 1 to $N$ and no retailer connected to more than one distributor, so the resulting graph is a tree rooted at the supplier in which the supplier links to all distributors and each distributor links to its own subset of retailers. The transport lead time and transport cost for each connection are sampled uniformly. (Generated SCN example: an infinite supplier, two distributors ($D_1$, $D_2$), four retailers, the distributor nodes replenish from the supplier, the first two retailers replenish from $D_1$, and the rest from $D_2$.)

Most real SCs are stochastic, so the deterministic demand setting studied here is a deliberate simplification.
It removes the need for multiple replications per data point, whereas stochastic demand would make each target a mean over replications, adding label noise and a proportional increase in generation cost.
Table~\ref{tab:param_ranges} summarizes the ranges used for all randomized parameters in the dataset.

\begin{table}[h]
    \centering
    \caption{Ranges of randomized parameters in the dataset. $\mathcal{U}\{a,b\}$ and $\mathcal{U}[a,b]$ denote discrete and continuous uniform distributions. All quantities are in arbitrary units.}
    \label{tab:param_ranges}
    \footnotesize
    \begin{tabular}{lll}
        \hline
        \textbf{Parameter} & \textbf{Applies to} & \textbf{Range} \\
        \hline
        No.\ of distributors $K$         & Network topology     & $\mathcal{U}\{1, 5\}$ \\
        No.\ of retailers $N$            & Network topology     & $\mathcal{U}\{K, 10\}$ \\
        No.\ of retailer connections $M$ & Per distributor      & $\mathcal{U}\{1, N\}$ \\
        Inventory capacity (S)           & Distributors, retailers & $S_D=\mathcal{U}\{200, 1000\}$, $S_R=\mathcal{U}\{10, 200\}$ \\
        Reorder level (s)                & Distributors, retailers & $s_D=\mathcal{U}\{200, S_D\}$, $s_R=\mathcal{U}\{10, S_R\}$ \\
        Daily demand arrival rate        & Retailers            & $\mathcal{U}\{1, 30\}$ \\
        Inventory holding cost           & Distributors, retailers & $h_D=\mathcal{U}[0.1, 0.6]$, $h_R=\mathcal{U}[0.4, 1]$ \\
        Product sell price               & Retailers            & $100 + \mathcal{U}\{1, 10\}$ \\
        Transport lead time (days)       & Edges                & $\mathcal{U}\{1, 10\}$ \\
        Transport cost                   & Edges                & $\mathcal{U}\{1, 100\}$ \\
        \hline
    \end{tabular}
\end{table}

A single simulation run takes approximately 6 seconds on a personal computer (13th-generation Intel Core i3 with 5 cores, 8~GB RAM, running Windows 11), meaning that generating the full dataset with 100,000 data points sequentially would require approximately 7 days.
To reduce this to a practical duration, we use Python's \texttt{multiprocessing} package to parallelize the independent data points generation across a server with two 20-core Intel Xeon Gold 6148 CPUs at 2.40~GHz (40 physical cores, 80 logical processors, running Linux), achieving an 80$\times$ speedup and reducing total generation time to approximately 2 hours.

SC structure and context can vary along several dimensions, namely the number of echelons, topology density and connectivity, the replenishment policy, product perishability, the stochasticity of demand and lead times, and node and link disruptions.
This dataset deliberately fixes a two-echelon, deterministic, non-perishable, min-max replenishment policy subset of this space, while SupplyNetPy supports the broader space for further datasets. 
\section{Exploring GNN Metamodel Architectures}
\label{sec:gnn_arch}

This section addresses Q1 and Q3 from the introduction: what GNN architecture is suited for SC performance regression, and does it generalize to unseen networks?
We describe the design of a custom message-passing GNN and present accuracy and generalization results.

\subsection{Problem Setup}
\label{subsec:problem_setup}

We predict the five node-level performance metrics (inventory carry cost, inventory spend cost, transport cost, revenue, and profit) from the SC graph structure and input parameters.
Each node carries five continuous input parameters, namely the inventory capacity (order-up-to level) $S$, the reorder level $s$, the inventory holding cost, the daily demand rate, and the product sell price, with parameters that do not apply to a node type, such as the demand rate and sell price for non-retailers, set to zero. Each edge carries two parameters, namely the transport cost and the transport lead time, and a learned indicator distinguishes the three node types (supplier, distributor, retailer).
The present study fixes the replenishment policy to min-max $(s,S)$ and varies only these continuous parameters together with the network structure. 
Categorical and structural factors such as the replenishment-policy type, the supplier-selection rule, node failure probability, and product perishability are held out and left to future work. This work therefore establishes the applicability of GNN metamodels to this class of SC problems rather than exhausting it.

\subsection{Architecture}
\label{subsec:architecture}

\textit{Architectural rationale.}
Four requirements specific to the regression task shape the architecture.
First, edge attributes such as transport cost and lead time directly determine node costs and revenue, so they must enter the message as vectors, unlike the scalar-weight GCN and GraphSAGE layers, motivating the custom message-passing layer of Equation~\ref{eq:msg_sc}.
Second, a node's performance depends on both upstream suppliers and downstream customers, so on the supplier-to-retailer directed graph we add back edges with a direction flag to restore downstream-to-upstream influence.
Third, since suppliers, distributors, and retailers play distinct roles, a learned node-type embedding lets the layer specialize by role.
Fourth, several node metrics are additive over a node's connections, for example a distributor's throughput sums the demand of the retailers it serves, so we use sum aggregation, which together with a single message-passing layer local to each node's immediate neighborhood keeps per-node prediction semantics independent of network size and is the main reason a single trained model transfers to much larger networks. Section~\ref{subsec:ablation} isolates the contribution of each choice.

\textit{Custom message-passing layer.}
GNNs operate on a graph $G = (V, E)$ where each node $v \in V$ carries a $d_n$-dimensional feature vector $\mathbf{x}_v$ and each edge $(u,v) \in E$ carries a $d_e$-dimensional feature vector $\mathbf{e}_{uv}$.
At each layer $l$, a node aggregates messages from its neighbors $\mathcal{N}(v)$ via MSG and AGG functions to update its hidden representation $\mathbf{h}_v^{(l)}$, initialized as $\mathbf{h}_v^{(0)} = \mathbf{x}_v$.
Our custom message-passing layer, an instance of the MPNN framework~\shortcite{gilmer2017neural}, computes the message from neighbor $u$ to node $v$ from the concatenation of node representations and the edge feature vector, followed by a learned transformation:
\begin{equation}
    \mathbf{m}_v^{(l)} = \text{AGG}\left(\left\{ \text{MSG} \left( \text{CONCAT} \left(\mathbf{h}_v^{(l-1)},\, \mathbf{h}_u^{(l-1)},\, \mathbf{e}_{uv} \right) \right) : u \in \mathcal{N}(v) \right\}\right)
    \label{eq:msg_sc}
\end{equation}
Here MSG is an $L$-layer ReLU MLP applied to $\mathbf{z} = \text{CONCAT}(\mathbf{h}_v^{(l-1)},\, \mathbf{h}_u^{(l-1)},\, \mathbf{e}_{uv})$, mapping from dimension $2d_n + d_e$ through $d_h$ hidden units to output dimension $d_{out}$, AGG sums the resulting messages, and UPDATE applies a ReLU to the summed message $\mathbf{m}_v^{(l)}$ to produce $\mathbf{h}_v^{(l)}$.

\textit{Implementation details.}
We add a back edge for each forward edge and append a direction flag to every edge, namely $+1$ for forward edges and $-1$ for back edges.
All features and outputs are min-max normalized.
Node type (supplier, distributor, retailer) is encoded via a learned embedding (\texttt{nn.Embedding}) and concatenated to the node feature vector.
An encoder MLP projects node and edge features into 32- and 16-dimensional latent spaces before the message-passing layer, and a decoder MLP maps the output to performance metric predictions.
The model is trained with Adam and mean squared error (MSE) loss.

\subsection{Experimental Setup}
\label{subsec:experimental_setup}

\textit{Datasets.}
All models are trained on the full dataset of Section~\ref{sec:data}, split into 80\% training and 20\% test sets.
Two further datasets, disjoint from the training data, serve as independent held-out evaluation.
The first is an unseen set of moderately larger networks generated with the same SupplyNetPy script and the same parameter ranges, sampling the number of distributors $K$ uniformly over 10 to 15 and the number of retailers uniformly over $K$ to 50, compared to 1 to 5 and $K$ to 10 in the training data.
The second is a scaling set of networks ranging from 21 to 421 nodes, from just beyond the largest training network (at most 16 nodes) up to more than twenty times its size, generated under two regimes. In the first regime the networks retain the training parameter ranges, isolating extrapolation in network size and topology, while in the second the simulation parameter ranges are additionally shifted to disjoint intervals, with node capacities about ten times larger, daily demand drawn from 50 to 100 rather than 1 to 30, and unit transport cost from 100 to 200 rather than 1 to 100. Each network size is evaluated on 100 generated graphs.

\textit{Training configuration.}
We select the hyperparameters on the full dataset by limited manual exploration rather than by optimization against the test set.
A single GNN layer suffices, since each node's performance depends primarily on its immediate neighbors.
Among hidden sizes 64, 128, and 256, 128 balances accuracy and training time.
A fixed learning rate of 0.001 yields stable convergence.
Training and test errors saturate by 30 to 40 epochs. We train for 100.
The message-passing MLP uses five linear-ReLU layers.

\textit{Evaluation metrics.}
We assess predictions with three metrics. The coefficient of determination $R^2$ has a maximum of 1 for perfect prediction, equals 0 for a predictor no better than the mean, and is negative for worse predictors.
To characterize the tail of the error distribution we additionally report the mean absolute error (MAE) and the 95th-percentile absolute error (P95).
Because all targets are min-max normalized to the unit interval, both MAE and P95 are absolute errors expressed in the normalized output space.

\subsection{Results}
\label{subsec:results}

\begin{figure}
    \centering
    \includegraphics[width=0.40\textwidth]{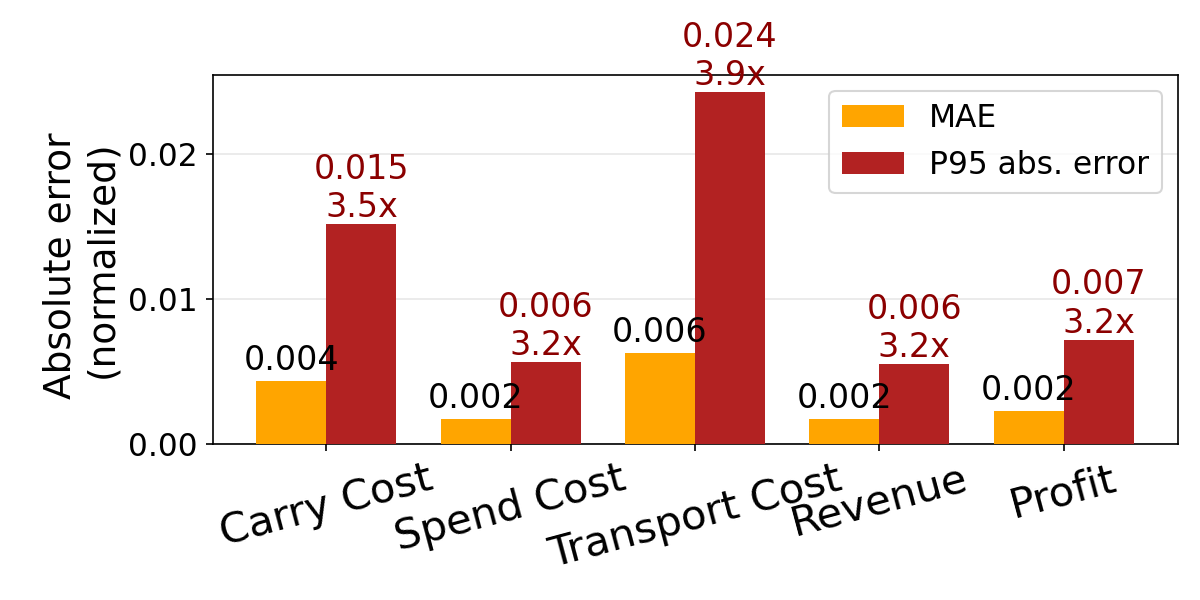}
    \includegraphics[width=0.40\textwidth]{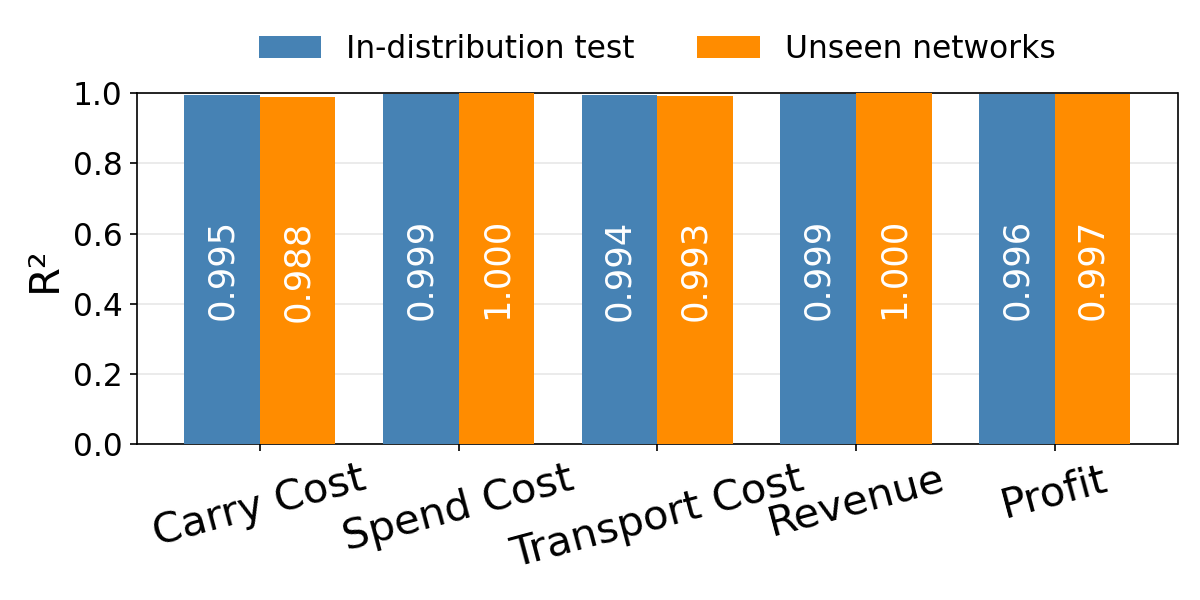}
    \caption{(a) per-metric MAE and P95 in the normalized $[0,1]$ output space. (b) per-metric $R^2$ on the held-out test set and the unseen larger networks, both from the single model trained on the training split.}
    \label{fig:worst_case}
\end{figure}

Under the final configuration and trained on the 80\% training split of the full dataset, the model achieves a training MSE of $3.31 \times 10^{-6}$ and a held-out test MSE of $1.93 \times 10^{-6}$.
Figure~\ref{fig:worst_case} (b) reports the $R^2$ scores for five node-level performance metrics on the held-out test set.
The model achieves $R^2 \approx 0.99$ across all metrics, so the figures display only the positive 0 to 1 range in which the results lie.
The simulation that generates the targets is the ground-truth oracle at $R^2 = 1$.
The high $R^2$ and the small gap between training and test errors are expected, because the simulation targets are deterministic and noise-free and the dataset is large relative to the small model.
The test error even falls marginally below the training error, which is expected here because the min-max normalization range is fixed from the training split, so the most extreme and hardest-to-fit target values lie in the training set rather than the held-out set.
The effect of training data size and MLP depth on $R^2$ is analyzed in Section~\ref{sec:tradeoffs}.
Network-level performance metrics can be obtained by pooling node-level predictions across all nodes in the network without any additional computation.

\textit{Worst-case error analysis.}
$R^2$ summarizes average predictive accuracy but can mask tail errors, which matter more than average errors when a surrogate is queried at the configurations an optimizer believes are promising.
We report per-metric MAE and P95 on the held-out test set, for the model trained on the full dataset with five MLP layers.
As shown in Figure~\ref{fig:worst_case}, P95 ranges from roughly 3$\times$ to 5$\times$ the MAE depending on the metric.
In physical units, de-normalizing by each target's range gives a test-set MAE of about 4.8 for profit and 0.63 for transport cost, under 1\% of their ranges ($-555$ to $1546$ and $0$ to $100$). As the simulation generates these targets, this is exactly the GNN's deviation from it.
Transport cost exhibits the heaviest tail, consistent with its comparatively lower $R^2$ score among the five metrics.

\textit{Comparing architecture variants.}
\label{subsec:ablation}
Having shown that the full model attains $R^2 \approx 0.99$ across metrics, we isolate the effect of each architectural choice from Section~\ref{subsec:architecture} by training reduced variants from scratch under one identical protocol (the same 80\% training split of the full dataset, min-max normalization, hidden size, learning rate, and epoch budget) and report the per-metric $R^2$ on the unseen larger-network set (Table~\ref{tab:ablation}).
These variants are trained together in a run separate from the main accuracy results, so the full model here reproduces the main model up to small run-to-run variation from the independently sampled split and weight initialization rather than reporting identical numbers.
We hold all hyperparameters fixed across variants rather than retuning each, and use no per-variant validation split because no per-variant model selection is performed, so the differences isolate the architectural effect rather than tuning effort.
The resulting gaps are representational rather than tuning artifacts, since a layer that cannot observe downstream demand or ingest edge attributes lacks information no hyperparameter setting can supply.
Only the full model and the variant without the node-type embedding keep every metric at $R^2$ above 0.98. The other variants fall to $R^2$ between roughly 0.5 and 0.9 on at least one metric, where predictions deviate visibly from the simulator and the surrogate is no longer reliable.
The back edges matter most, since removing them collapses the metrics that depend on downstream demand, namely revenue, spend cost, and profit, to below 0.14 once distributors can no longer observe their retailers, even though carry and transport cost survive.
The vector-valued edge features are essential for transport cost, whose $R^2$ falls to about 0.52 in every variant that cannot ingest them.
The node-type embedding contributes the least, because distributors and retailers share near-identical inventory dynamics here, and it is expected to matter when genuinely heterogeneous node types are present.
The standard GCN and GraphSAGE layers, run here on the raw node features without the custom encoders or node-type embedding, exceed a mean predictor but stay well short of a usable surrogate, trailing the custom architecture by a wide margin. The single-factor rows pinpoint the decisive components, namely the edge-conditioned message and the back edges.

\begin{table}[h]
    \centering
    \caption{Per-metric $R^2$ of each architecture variant on the unseen larger-network set. All variants are trained from scratch under one identical protocol.}
    \label{tab:ablation}
    \footnotesize
    \begin{tabular}{lccccc}
        \hline
        \textbf{Variant} & \textbf{Carry} & \textbf{Spend} & \textbf{Transport} & \textbf{Revenue} & \textbf{Profit} \\
        \hline
        GCN (scalar edge weight)             & 0.807 & 0.860 & 0.532 & 0.862 & 0.818 \\
        GraphSAGE (no edge features)         & 0.946 & 0.866 & 0.515 & 0.868 & 0.877 \\
        Custom, no edge features             & 0.951 & 0.962 & 0.518 & 0.964 & 0.966 \\
        Custom, no node-type embedding       & 0.989 & 0.999 & 0.987 & 0.999 & 0.997 \\
        Custom, no back edges                & 0.946 & 0.138 & 0.877 & 0.137 & 0.103 \\
        Custom (full)                        & 0.989 & 0.999 & 0.989 & 0.999 & 0.997 \\
        \hline
    \end{tabular}
\end{table}

\begin{figure}
    \centering
    \includegraphics[width=0.42\textwidth]{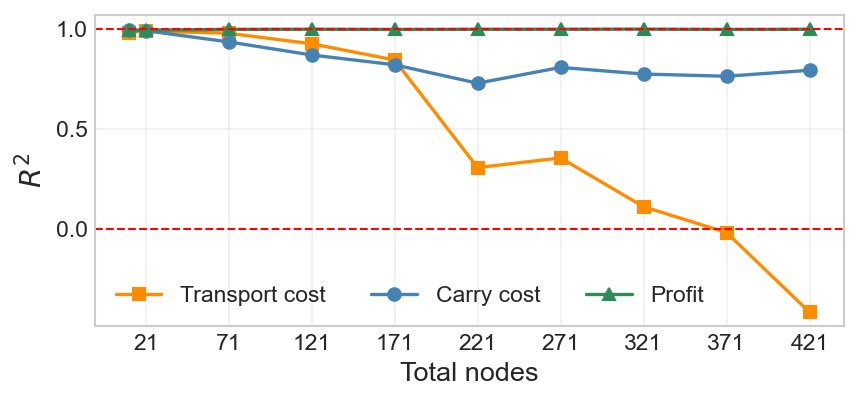}
    \includegraphics[width=0.42\textwidth]{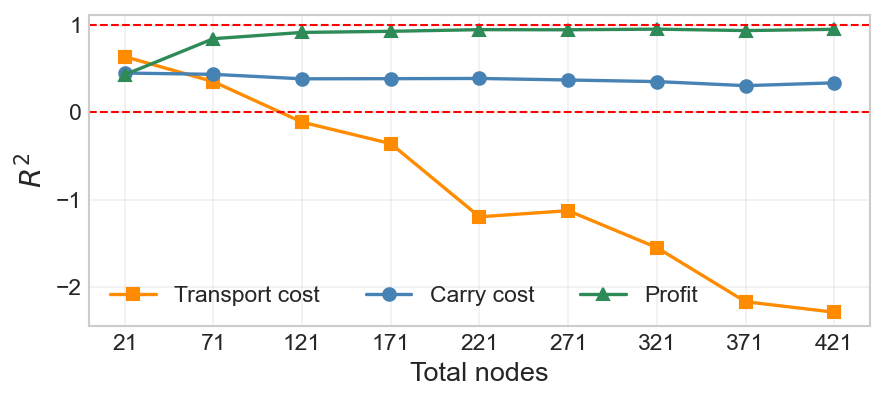}
    \caption{$R^2$ of profit, transport cost, and inventory carry cost versus network size, under (a) training parameter ranges and (b) shifted parameter ranges. Dashed and dotted lines mark $R^2 = 1$ and $0$.}
    \label{fig:gen_scaling}
\end{figure}

\textit{Generalization to unseen networks.}
As a first check, we evaluate the trained model on the unseen set of moderately larger networks defined in Section~\ref{subsec:experimental_setup}.
Figure~\ref{fig:worst_case} (b) reports the $R^2$ scores on these networks alongside the test-set scores, and the two are comparable, indicating that the GNN metamodel generalizes to larger networks rather than memorizing the training graphs.
To probe generalization more rigorously, we evaluate the same single trained model on the scaling set of Section~\ref{subsec:experimental_setup}, whose networks range from 21 to 421 nodes, all larger than the largest training network, under two regimes, one retaining the training parameter ranges and one shifting them to disjoint intervals.
Profit, spend cost, and revenue stay near 0.99 across the full size range under the training ranges, and spend cost and revenue remain near 0.99 under the shift, so the metamodel extrapolates well to networks far larger than those seen in training.
Profit stays accurate on the largest shifted networks but scores lower on the smaller ones, where its values vary little and the variance-normalized $R^2$ falls (Figure~\ref{fig:gen_scaling}).
Inventory carry cost and transport cost mark the boundary of the metamodel's valid region (Figure~\ref{fig:gen_scaling}), with carry cost degrading from about 0.99 to roughly 0.8 at the largest networks under the training ranges and to about 0.4 under the shift, reflecting time-averaged inventory driven by the replenishment policy, demand, and lead time.
Transport cost degrades the most, its $R^2$ turning negative beyond about 350 nodes under the training ranges and reaching roughly $-2.3$ on the largest shifted networks.
Extending the metamodel to stay accurate for transport and carry cost on much larger and parameter-shifted networks is left to future work.

\section{Computational Trade-offs}
\label{sec:tradeoffs}

This section addresses Q4 from the introduction: are GNN metamodels computationally efficient relative to running the simulation model directly?
Two sources of computational cost are relevant when using a GNN metamodel for SBO, namely the one-time cost of generating training data and training the model, and the per-evaluation cost of a single forward pass through the trained model at inference time.
We analyze the first through the joint trade-off between prediction accuracy, training-set size, and architecture complexity, together with the resulting training time, and then compare the per-evaluation forward-pass cost against that of a full simulation run.

\subsection{Effect of Training Data Size and Architecture Complexity on Prediction Accuracy}
\label{subsec:data_size_arch}

We study how training-set size $D$ and message-passing MLP depth $L$ govern accuracy, sampling $D \in \{10{,}000,\dots,80{,}000\}$ from the full dataset (seed 42) and training one model per $(D, L)$ pair with $L \in \{2,\dots,7\}$, all other hyperparameters held at their final values (Section~\ref{subsec:experimental_setup}). We report $R^2$ for profit, which aggregates a node's cost and revenue components and thus reflects overall SC performance. Even with 10,000 points the model is already accurate, but extra depth does not help at small $D$, where capacity is underused. Accuracy improves with $D$ and saturates beyond roughly 60,000 points, after which it also rises with $L$ up to a stable region. A user can therefore trade accuracy against data-generation and training budget by choosing $(D, L)$, configurations that Figure~\ref{fig:forward_vs_sim} places on a common axis of accuracy and per-evaluation cost. The final configuration uses five layers, a small margin above this minimum within the stable region, with the full dataset for the main results (Section~\ref{sec:gnn_arch}).

\subsection{Training Time}
\label{subsec:training_time}

We measure training time over the same sweep of training-set size $D$ and MLP depth $L$.
Training time is measured using Python's \texttt{time} library on the server described in Section~\ref{sec:data} (Intel Xeon Gold 6148), with each model trained on a single core under the CPU build of PyTorch.
Training time scales approximately linearly with both dataset size and MLP depth, providing a reliable reference for estimating training cost.
Across the swept configurations, single-model training ranges from about 6~minutes for the smallest setting (10,000 points, 2 layers) to about 54~minutes for the largest (80,000 points, 7 layers).

\subsection{GNN Forward Pass versus Simulation Run}
\label{subsec:forward_vs_sim}

\begin{figure}[htb]
    \centering
    \includegraphics[width=0.8\textwidth]{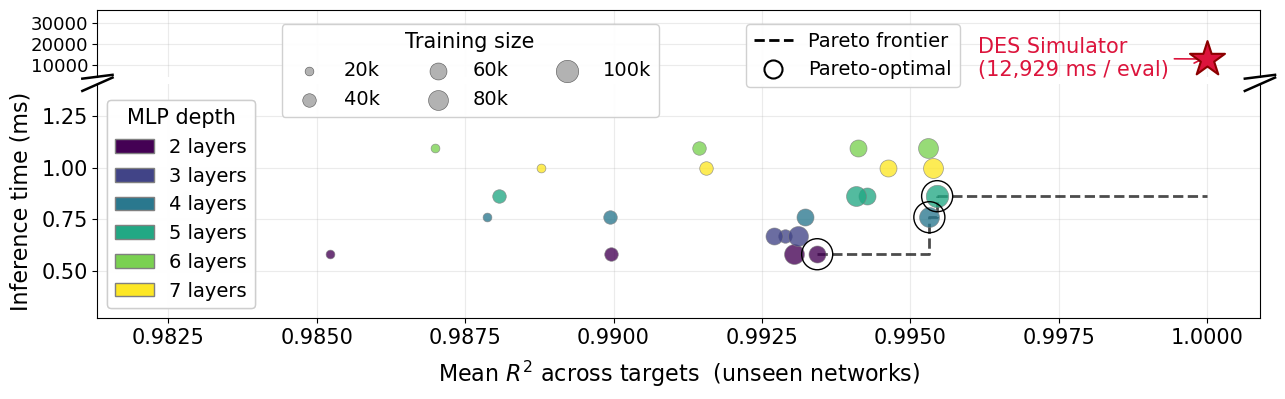}
    \caption{Accuracy ($R^2$) versus computational cost (per-evaluation time) for the GNN metamodel and the simulation model.}
    \label{fig:forward_vs_sim}
\end{figure}

Once trained, the GNN metamodel can replace the simulation model in the optimization loop.
On the personal computer, a single simulation run used to generate one data point takes approximately 6~seconds, whereas a single GNN forward pass over one network completes in about 1~millisecond, more than three orders of magnitude (roughly 6000 times) faster.
Figure~\ref{fig:forward_vs_sim} compares the accuracy ($R^2$ score) and computational cost (GNN forward pass time versus simulation run time) of the GNN metamodels (with different depths and training data sizes) and the simulation model, with both timed on unseen networks larger than those used for training, where a single simulation run takes about 13~seconds.
Because an optimizer must evaluate the model at many different configurations, replacing the simulation with the trained GNN metamodel allows this search to proceed at a fraction of the computational cost, making the GNN a practical surrogate for SC performance prediction.
Moreover, the forward-pass time stays close to 1~millisecond for networks ranging from roughly 20 to 70 nodes, while simulation time grows with both network size and simulation length, so the metamodel's advantage over simulation only widens as networks grow.

\section{Conclusion and Future Directions}
\label{sec:conclusion}

The application of GNNs to SCs is still sparse, and to the best of our knowledge, no prior work has demonstrated their use as a simulation surrogate that learns to predict node- and network-level performance and to generalize over arbitrary, previously unseen SC graphs. This paper is the first to present findings of an exploration into this direction.
We revisit the five research questions raised in the introduction. Q1, Q3, and Q4 are addressed by the empirical results, Q2 by the dataset contribution, and Q5 defines the research agenda that this paper sets out and that we are pursuing.

\textit{Q1: can GNNs work well as surrogates for SC performance prediction, and what architectures are suitable?} We presented a custom message-passing architecture that incorporates both node and edge features, and showed that it predicts five node-level performance metrics (inventory carry cost, inventory spend cost, transport cost, revenue, and profit) with $R^2 \approx 0.99$ on held-out data.

\textit{Q2: how can we generate training data for the GNN metamodel?} This is met by the publicly released dataset of 100,000 programmatically generated two-echelon SCNs with randomized topology and parameters, labeled with steady-state performance metrics obtained via DES using SupplyNetPy. To our knowledge, it is the first open resource of its kind.

\textit{Q3: do such surrogates generalize to SC graphs not seen during training?} On moderately larger unseen networks the model maintains $R^2 > 0.98$ across all metrics. Pushed to networks more than twenty times the training scale, and to shifted parameter ranges, profit, revenue, and spend cost largely retain their accuracy, whereas inventory carry cost and transport cost degrade and mark a validity boundary. The GNN therefore learns structural relationships rather than memorizing specific network sizes.

\textit{Q4: is the surrogate computationally attractive relative to simulation?} The accuracy-compute trade-off differs in kind between the two, since GNN cost and accuracy depend on architecture, depth, hidden size, and training-set size, whereas simulation cost and accuracy depend on inherent variance, the number of replications, and the simulated time horizon. Our preliminary experiments confirm a clear advantage: at comparable predictive accuracy, a single GNN forward pass is orders of magnitude faster than a full simulation run, making it a drop-in for use in an optimization or design-space exploration loop.

\textit{Q5: can GNNs' differentiability be leveraged for optimization and design-space exploration?} This is not demonstrated empirically here and is the forward-looking question that defines our research agenda. A trained GNN metamodel is a differentiable function of its inputs that interpolates continuously over them. What is distinctive is that a GNN takes node parameters, edge parameters, and graph structure jointly as input, so its continuity covers graph structure and not only the parameter values that most neural surrogates handle. Our generalization to unseen graphs supports this, indicating the learned function is smooth not only across parameters at a fixed topology but also across topologies themselves. This opens a pathway that classical metamodels cannot support, namely joint parameter and structural optimization of an SC via gradient-based methods together with fast what-if analysis over both policies and topologies.

\subsection{Future Work}
\label{sec:optimization}
The work reported here is an initial step, and the directions ahead form a single connected arc from broadening the data to closing the optimization loop. 
The most immediate need is to enrich the training dataset beyond its current two-echelon, single-product, deterministic-demand scope, adding greater echelon depth, heterogeneous node and edge types, stochastic demand, richer replenishment policies, and broader parameter ranges. 
A richer dataset in turn motivates a more systematic architecture study, since identifying a single architecture that works does not reveal which classes work and why; mapping families such as GCN, GAT, GIN, NNConv, GINE, PNA, and heterogeneous variants would expose the inductive biases that matter for SC structure and sharpen the accuracy-compute Pareto frontier. 
That frontier is also where the cost case must be made rigorous, by characterizing the smallest architecture that achieves a target accuracy as a function of graph depth and complexity and plotting its accuracy-compute curve against the simulator's own, turning the present preliminary comparison into a principled statement of when the surrogate is preferable. 
The payoff of all three, and the ultimate aim of this line of work, is to use the differentiable metamodel for SC optimization and design-space exploration, namely formulating the joint parameter-and-topology optimization problem (for example, via continuous relaxation of edge-existence probabilities), performing sensitivity analysis through the trained metamodel, and developing gradient-based methods that exploit differentiability to descend toward improved designs. This is the novel and rich research landscape that the present paper aims to open up, and it is the focus of our ongoing work.

\bibliographystyle{plainnat}
\bibliography{bibliography}

\end{document}